\documentclass{llncs}
\usepackage{cite}
\usepackage{epsf}
\usepackage [english]{babel}
\usepackage [autostyle, english = american]{csquotes}
\MakeOuterQuote{"}

\usepackage{tikz}

\tikzstyle{box} = [rectangle, minimum width={width("Get Synonyms and Metaphors")+0.2em}, minimum height=2em, text centered, draw=black]

\tikzstyle{io_box} = [trapezium, trapezium left angle=70, trapezium right angle=110, minimum width={width("Get Synonyms and Metaphors")+0.2em}, minimum height=2em, text centered, draw=black]

\tikzstyle{arrow} = [thick,->,>=stealth]
\tikzstyle{every node}=[font=\small]
\usetikzlibrary{shapes.geometric, arrows}
\usepackage{tikz}
\tikzstyle{box} = [rectangle, minimum width={width("Get Synonyms")+0.2em}, minimum height=2em, text centered, draw=black]
\tikzstyle{io_box} = [trapezium, trapezium left angle=70, trapezium right angle=110, minimum width={width("Get Synonyms")+0.2em}, minimum height=2em, text centered, draw=black]
\tikzstyle{arrow} = [thick,->,>=stealth]
\tikzstyle{every node}=[font=\small]
\usetikzlibrary{shapes.geometric, arrows}

\usepackage{url}
\urldef{\mailsa}\path|ghiranan@adobe.com,pmanerik@adobe.com,jharsh@cs.cmu.edu|

\begin{document}

\title{Generating Appealing Brand Names}

\author{Gaurush Hiranandani\inst{1} \and Pranav Maneriker\inst{1} \and 	Harsh Jhamtani\inst{2} }

\institute{Adobe Research, Bangalore, India \and
Language Technology Institute, Carnegie Mellon University, USA
\mailsa
}

\maketitle

\begin{abstract}

Providing appealing brand names to newly launched products, newly formed companies or for renaming existing companies is highly important as it can play a crucial role in deciding its success or failure. In this work, we propose a computational method to generate appealing brand names based on the description of such entities. We use quantitative scores for readability, pronounceability, memorability and uniqueness of the generated names to rank order them. A set of diverse appealing names is recommended to the user for the brand naming task. Experimental results show that the names generated by our approach are more appealing than names which prior approaches and recruited humans could come up. 

\end{abstract}

\section{Introduction}
\label{sec:introduction}

Choosing right brand names for newly launched products, newly formed companies and entities like social media campaigns, apps, websites etc. is critical. In the context of creating brands, it is believed that such a naming decision may well be "the most important marketing decision one can make"~\cite{Ries:81}. A marketer may often spend a lot of time in coming up with an appealing name which can achieve favorable outcomes on various key performance indicators (KPIs) like website visits, number of customer acquisitions, etc. This becomes critical in scenarios like quickly planned campaigns, where there is not enough time for marketers or authors to come up with an appealing name. 
However, prior technologies are insufficient to computationally come up with appealing names for such entities based on a provided description. Moreover, rarely is the management provided with interpretable objective criteria upon which a brand name is suggested~\cite{Ries:81}. 
This creates a need for an algorithm which automatically generates appealing names from the description of an entity in a justified manner.

This paper has the following contributions. Firstly,  we define and infer the importance of various linguistic and statistical features for the task of suggesting names for brands, products or other such entities. 
Secondly, we propose computational methods to generate brand names given the description of the entity in question. Though coming up with names for entities like brands is considered mostly a creative task, our MTurk based evaluation study determining the appeal of the recommended names shows that names from our method obtain ratings comparable with human-provided names.

Rest of the paper is organized as follows. Section \ref{sec:reviewofliterature} reviews the existing relevant works. Section \ref{sec:methodology} explains methodology behind generating and ranking names. Section \ref{sec:evaluation} explains the conducted experimental studies and evaluation. In Section \ref{sec:limitations} we discuss some of the limitations and future work. Lastly in Section \ref{sec:conclusions}, we provide conclusions.

\section{Related Work}
\label{sec:reviewofliterature}
Robertson \cite{Robertson:89} showed characteristics of a `good' name which include short, easy to say, spell, read, understand and easily retrievable from memory. Yorkston and Menon \cite{Yorkston:04}
showed consumers use the information they gather from phonemes in brand names to infer product's attributes. Little et al. \cite{Little:10} suggested that a recommendation tool for performing the naming task better should aid in ideation as well.
 
From the perspective of word-generation, there are prior works on password and domain name generation.
Some studies have focused on memorability of passwords~\cite{Clements:15}, while tools like \textit{PWGEN}~\cite{Allbery:88} and \textit{Kwyjibo}~\cite{Crawford:08} generate pronounceable passwords and domain names respectively.
However, using such tools and being limited to one attribute is not useful for brand-naming generation given the description.

Bauer \cite{Bauer:83} treated blending of words\footnote{forming a word by combining sounds from two or more distinct words - e.g. \textit{Wikipedia} by blending ``Wiki'' and ``encyclopedia''} as a process to create neologisms.
Such blending can be based on various phonetic and syllable alignment techniques, such as those used by Kondrak \cite{kondrak2003phonetic} and Hedlund et al. \cite{hedlund2005natural}.
{\"O}zbal and Strapparava \cite{ozbal:12}
proposed a computational approach to generate neologisms consisting of homophonic puns and metaphors based on the category and properties of the entity.
However, for recommendation purposes, it is important to define a ranking based on appeal of a name given the properties or description of the entity. {\"O}zbal and Strapparava \cite{ozbal:12} carry out filtering/ranking during the evaluation by combining the phonetic structure and language model with equal weights.
However, ranking based on the appeal of a name is not motivated by the crucial metrics mentioned above. This becomes critical when one needs to recommend a few appealing names rather than generate a large number of names.
Further, some online tools\footnote{www.online-generator.com, www.namegenerator.biz} provide names by concatenating random strings and some consulting companies\footnote{ABC Namebank, A Hundred Monkeys} are engaged in brand naming but do not use any automated processes.

\section{Methodology}
\label{sec:methodology}
\begin{figure*}[!h]
\centering
\begin{tikzpicture}

	\node (step1) [io_box, align=left] {Take input\\ as Description};
    \node (step2) [box, right of=step1, xshift=8em] {Remove Stop Words};
    \node (step3) [box, right of=step2, xshift=8em] {Find POS Tags};
    \node (step4) [box, below of=step3, yshift=-1em, align=left] {Get Synonyms\\ and Metaphors};
    \node (step5) [box, left of=step4, align=left, xshift=-8em] {Break Words\\ in Syllables};
    \node (step6) [box, left of=step5, xshift=-8em] {Generate Blend Names};
    \node (step7) [box, below of=step6, yshift=-1em] {Give Appeal Scores};
    \node (step8) [box, right of=step7, xshift=8em] {Rank Names};
    \node (step9) [io_box, right of=step8, xshift=8em, align=center] {Recommend Diverse\\ Appealing Names};
    
    \draw [arrow] (step1) -- (step2);
    \draw [arrow] (step2) -- (step3);
    \draw [arrow] (step3) -- (step4);
    \draw [arrow] (step4) -- (step5);
    \draw [arrow] (step5) -- (step6);
    \draw [arrow] (step6) -- (step7);
    \draw [arrow] (step7) -- (step8);
    \draw [arrow] (step8) -- (step9);

\label{diag:workflow}    
\end{tikzpicture}
\caption{Overview of the algorithm}
\end{figure*}
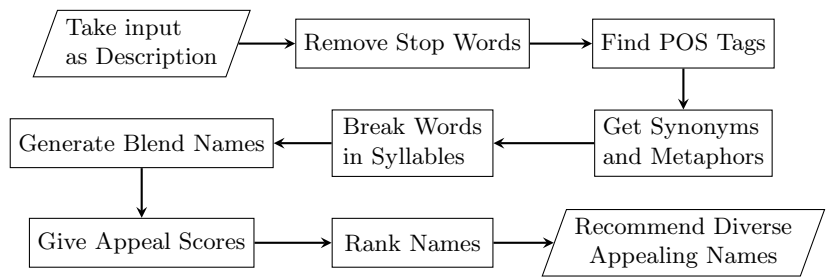

Figure \ref{diag:workflow} provides an outline of our proposed solution. We do some basic preprocessing on the words in description, followed by expansion of the set of words using an external ontology. We blend the words to generate candidate names. Then, we score and rank the names based on readability, memorability, pronounceability and uniqueness. Finally we postprocess the ranked list of names to provide a diverse set of suggestions.

As mentioned earlier, we generate candidate names based on blending of syllables present in the word set. The choice of following this approach is based on prior works.
 {\"O}zbal et al. \cite{ozbal:12:brand} provided an annotated dataset of 1000 brand names to understand linguistic creativity involved in naming.
In the data, around 20\% of the names are created either by juxtaposition or clipping which are morphological mechanisms similar to blending.\footnote{e.g. \textit{DocuSign} from ``Document'' and ``Signature''} Further, Bauer \cite{Bauer:83} treated blending of words as a process to create neologisms.
This convinced us to
generate candidate names based on blending of words. 


\subsection{Generate Names}
\label{ssec:generatenames}  

%
%
%
%

The method takes the description of an entity as input. 
For example, an input can be `\emph{\textbf{Creating} an \textbf{application} to \textbf{split} \textbf{expense} \textbf{wisely}}.'
We use words other than stop-words~\cite{Bird:09} while generating names and call them \emph{root} words.
Thereafter, our method looks for part-of-speech (POS) tags~\cite{Toutanova:03} of the root words.
POS tags are important as words can have a different semantic orientation based on usage and thus have different set of synonyms (used later).
Let $D$ denote the set of words along with their POS tags. For example, 

$D$ = \{(Creating, Verb), (Application, Noun), (Split, Verb), (Expense, Noun), (Wisely, Adverb)\}

Next, we use \textit{Wordnet}~\cite{Wordnet:95} to obtain synonyms of root words based on POS. Further, we obtain synonyms and metaphors by applying a strategy similar to {\"O}zbal and Strapparava \cite{ozbal:12}. We call these words \emph{related words} and attribute the same POS tag as their root word.
Let $C_1$, $C_2$,.., $A_1$, $A_2$,.., $S_1$, $S_2$,.., $E_2$, $E_2$,.., $W_1$, $W_2$,.. be the related words obtained for \textit{Creating}, \textit{Application}, \textit{Split}, \textit{Expense} and \textit{Wisely} respectively. Further let $R$ denote the list of related words with POS tags. Then, 

\emph{$R$ = \{(Creating, Verb), ($C_1$, Verb), ($C_2$, Verb),..., (Application, Noun), ($A_1$, Noun), ($A_2$, Noun),..., (Split, Verb), ($S_1$, Verb), ($S_2$, Verb),..., (Expense, Noun), ($E_1$, Noun), ($E_2$, Noun),..., (Wisely, Adverb), ($W_1$, Adverb), ($W_2$, Adverb),...\}}
%

Each word in $R$ is split into syllables using PyHyphen.\footnote{https://pypi.python.org/pypi/PyHyphen/} We attach the same POS tag for each syllable as its parent word. Let us denote the set of syllables of the root and related words along with their POS tags by $L$. For example, 

$L$ = \{(Cre, Verb), (At, Verb), (Ing, Verb), (App, Noun), (Li, Noun), (Ca, Noun), (Tion, Noun),...\}

We observed certain \emph{rules} used in blended names from the description provided by the annotators in {\"O}zbal et al. \cite{ozbal:12:brand}. A \emph{rule} is a combination of unordered  POS tags of the blended words (syllables). For example, \emph{SplitWise} and \emph{WiseSplit} are created from two syllables whose POS tags are verb and adverb.
Hence, they follow the \emph{Verb-Adverb} rule.
Table \ref{blendresult} presents the percentages of each rule used in blended names.
 Note that the percentage of names created using a certain rule are subject to discretionary choice based on the annotator's bias. As in, for the same entity having similar properties, there can be two different descriptions from two different annotators leading to different empirical estimates. For example, \emph{data platform enables solving problems quickly} or \emph{data platform is quick in problem solving}. Here, we will have the same syllable \emph{quick} in both sentences. However, in the former case, the POS tag is Adverb whereas, in the latter, it is Adjective. For the results shown in Table~\ref{blendresult}, we have used the POS tags obtained from the description provided by the annotators in {\"O}zbal et al. \cite{ozbal:12:brand}.    
We observed that some of the blending rules were less frequent.  Hence, 
we omitted the rules having less than 1\% of the names under their category and called the rest as \emph{allowed rules}.
For example, a name using two syllables coming from words having verb as their POS tag is usually not created. An example can be \emph{SplitBreak} formed by the syllables \emph{Split} and \emph{Break}. 
With respect to the data, this essentially removed three rules but had a significant reduction in the number of names generated. 

\begin{table}[t]
	\small
	\centering
	\caption{\label{blendresult} Percentage of Blending Rules}	
	\begin{tabular}{llll}
		\hline \bf Rule & \bf \% & \bf Rule & \bf \% \\ \hline
		Noun-Adjective & 40.10 & Adjective-Adverb & 3.2 \\
		Noun-Verb & 8.02 & Noun-Noun & 36.36 \\
		Noun-Adverb & 4.81 & Verb-Verb & 0.00 \\
		Verb-Adjective & 0.53 & Adjective-Adjective & 3.28 \\
		Verb-Adverb & 3.7 & Adverb-Adverb & 0.00 \\
		\hline
	\end{tabular}
	
\end{table}

Finally, the method creates new blended names by joining two or three syllables at a time taken from the permutation set of $L$, given that the blending is within allowed rules. Let $N$ denote the set of generated names. 
For the given example, some names generated by two syllables are \emph{SplitWise}, \emph{BudSplit} 
and \emph{BreakOwl}. Note that \emph{break} is a synonym of \emph{split} and \emph{owl} is a metaphor for \emph{wise} obtained from the idiom - \emph{as wise as an owl}. One of the names generated by three syllables is \emph{ExPenseBreak}.
We prefer syllables over morphemes as the combination
of syllables can generate any name that a combination of morphemes can. The increased number of names generated due to this choice are handled by ranking and selecting the top candidate names. This is explained in the next section.


\subsection{Ranking Names}
\label{rankingnames}

\subsubsection{Scores Formulation.}
\label{ssec:scoresformulation}
Every description can potentially generate thousands of names using the above method. However, it is crucial to rank them for recommendation purposes. Therefore, each 
name is given a score based on the mathematical formulations of 4 features: readability, pronounceability, memorability, and uniqueness. The scores are normalized in the range of $[0, 1]$ over the English dictionary {\cite{Goldhahn:12}, with $0$ and $1$ representing the least and the maximum score respectively. Let $n$ and $|n|$ denote a name and its length respectively.
	
	\textbf{Readability:} A good name should be easy to read. Hence, to assign readability score, we use
	Flesch---Kincaid Reading-Ease Score~\cite{Kincaid:75}. Since each name is a single word, readability  (denoted by $R(.)$) becomes:
	\begin{equation}
	R(n) = 205.82 - 84.6*{|syllables(n)|} 
	\label{eq:read}
	\end{equation}
	
	Where, $|syllables(n)|$ denotes the number of syllables in $n$. Later in this section, we will observe formulation in Equation \ref{eq:read} being reduced to number of syllables due to a linear model computing appeal of a name.
	
	\textbf{Pronounceability:} The more permissible the combinations of phonemes is, the more pronounceable the word becomes. We adapt the concept from Schiavoni et al. \cite{Schiavoni:14} with some refitting to measure the extent with which a string adheres to the phonotactics of the language. By taking substrings (n-grams) of $n$ of length $l \in \{2,3,4\}$ with frequencies from the dictionary~\cite{Goldhahn:12}, we compute certain features as follows:
	\begin{equation}
	S_l^n = \frac{\sum \limits_{t\in n-grams(n)} freq(t)}{|n| - l + 1}
	\label{pronounce} 
	\end{equation}
	Here, $freq(t)$ is the frequency of the n-gram $t$ in the dictionary. 
	For example,
	$$
	S_2(facebook) = \frac{fa_{109} + ac_{343} + ce_{438} + eb_{29} + bo_{118}+ oo_{114} + ok_{109}}{8 - 2 + 1} = 170.8
	$$
	Feature values for smaller $l$ will be higher than larger $l$, but feature values for larger $l$ will be more important, since there are around 4000 meaningful 4-letter words in comparison to around 1000 meaningful 3-letter words in English~\cite{Mirriam:05}. Hence, the probability of a 4-letter word being meaningfully used will be higher than a 3-letter word. 
	Therefore, weights for $S_l^n$ are assumed as $w_l = \frac{l}{2+3+4}$ for $l = \{2,3,4\}$. Finally, we define pronounceability $P(.)$ as:
	\begin{equation}
	P(n) = w_2S_2^n + w_3S_3^n + w_4S_4^n
	\end{equation}
	This formulation is a simple back off model~\cite{chen1996empirical}, where we always back off to a lower order n-gram with fixed probability.
	
	
	\textbf{Memorability:} 
	Danescu-Niculescu-Mizil et al. \cite{Mizil:12} claim that memorable quotes use rare word choices, but at the same time are built upon a scaffolding of common syntactic patterns. Adapting it to blended syllables, we use Meaningful Characters Ratio defined by Schiavoni et al. \cite{Schiavoni:14} to capture memorability. It	
	models the ratio of characters of $n$ that comprise a meaningful word. A word is said to be meaningful if it occurs in~\cite{Mirriam:05}.
	We define memorability $M(.)$ as follows:
	\begin{equation}
	M(n) = \max_{all \, splits \, of \, n} \frac{\sum\limits_{i=1}\limits^k |s_i|}{|n|}
	\end{equation}

	Here, $s_i$ are the meaningful substrings of length $\geq 3$ obtained by splitting $n$ and $k$ is their number.
	For example, If $n = facebook$,
	$$
	M(facebook) = \frac{(|face| + |book|)}{8} = 1
	$$

	%
	%
	
	\textbf{Uniqueness:} This feature prefers names having low usage and non-dictionary words. Consider a time series $(V, T)$ for $n$ such that $V = \{v_1,... ,v_T\}$ and $T = \{t_1, ... , t_T\}$. $t_i's$ represent consecutive years
	and $v_i$ represents the normalized usage of $n$ in year $t_i$ as provided by GoogleNgrams~\cite{Michel:11}. Then, uniqueness $U(.)$ is defined as:
	\begin{equation}
	U(n) = \frac{\sum\limits_{k=1}\limits^{T_c} v_k*(t_k  - t_1)}{\sum\limits_{k=1}\limits^{T_c} (t_k  - t_1)}
	\end{equation}
	where $T_c$ represents the latest year. The intuition is that less usage in recent years is more important. Further, if GoogleNGrams fails to produce any time series for $n$, then $U(n)$ is taken to be 1. 
	
	
	
\subsubsection{Combining Scores.}
\label{ssec:combiningscores}
	
	We needed to model the appeal of a brand name depending on the quantitative definitions of the mentioned linguistic features. We asked few annotators to provide descriptions of the entities they want to name. Among the descriptions provided, we randomly chose three of them. Thereafter, we conducted a survey of 20 participants who were shown 3 lists of 15 names generated by our method, one list for each of the three descriptions. They were asked to rank the names in a list from 1 to 15. On an average, the Kendall-Tau correlation between the "average ranks" and the "individual ranks" came out to be 0.66, 0.68 and 0.62 for the 3 lists. High correlation suggested that people give similar rankings of names if shown a description. Hence, we took average ranks to be the ground truth rankings for the 3 lists.

Next we test if the four feature scores described earlier were correlated. In  other words, we wanted to test if the individual features provide information not covered by other features.
	While the pairwise Pearson's Correlation among the 4 scores based on 45 names ranged from $-0.48$ to $0.23$, the Kendall Tau correlation among ranking from each individual score ranged from $-0.48$ to $0.21$. This implied lack of correlation amongst the scores. Therefore, we defined appeal $A(.)$ of a name to be weighted linear combination of the mentioned scores. That is, 
	\begin{equation}
	A(n) = a_rR(n) + a_pP(n) +  a_{m}M(n) + a_uU(n)
	\end{equation}
	Where, $\vec{a} = (a_r, a_p, a_{m}, a_u)$ is the weight vector. 
	Then we applied rank-svm~\cite{Joachims:02} which used the obtained $315$ pairwise comparisons to learn the weights showing importance for different features. The learned weights were as follows: \\ $\vec{a} = (2.18, 1.63, 0.91, 1.05)$.
	Interestingly participants indicated \emph{readability} as the most crucial factor for comparing names amongst the four measures.
	For $n = SplitWise$, we obtained: $A(n) = 3.71$, $R(n) = 0.77$, $P(n) = 0.04$, $M(n) = 1.0$, $U(n) = 1.0$. 
	
	\subsection{Recommendation}
	\label{ssec:recommendation}
	
	The user can be recommended with top names as per their appeal scores but since a single syllable may appear in many top names (eg. \emph{fur}, \emph{con} etc. which are meaningful and easy to read), this set of recommendation may not help ideation. Hence, we diversify~\cite{carbonell1998use} the set of names to aid ideation by an update rule~\cite{modani2015creating}.
	The intuition is that after we choose the top candidate name based on appeal score, we update the appeal scores of the names formed by the same syllables as that of the top candidate name. And then choose the top name from the rest of the names having updated appeal scores. Suppose one chooses $n'$ in the first iteration, then the update for diversity is defined as:
	\begin{equation}
	A(n) \leftarrow \frac{1}{|m|*|k|}*A(n)  \quad \textrm{for} \quad  n \in N \setminus [\{n'\} \cup N']
	\label{eq}
	\end{equation}
	Here, $|m|$, $k$, and $N'$ denote the number of common syllables in $n$ and $n'$, number of syllables in $n$, and names sharing no common syllable with $n$ respectively.
	We iteratively choose the best candidate ($n'$) and then update appeal of other names ($n$) by using Equation \ref{eq}.
	The names chosen after some iterations (say 30) are recommended. This ensures that the recommended set of names as a whole become useful for the naming task. For the given example, the top 5 names are \emph{ConTear, BreakWise, BudSplit, BreakOwl} and \emph{DisCleave}.
	

	\section{Evaluation}
	\label{sec:evaluation}
	To estimate the quality of the generated names from our method, we conducted an MTurk study and compared our names with two baselines. 
	We created descriptions for 10 entities which usually require brand names. For example, one of the descriptions was \emph{Light-weight software to locate virus on computer}. Our method took 4 minutes on our machine to generate 984 ranked names on an average for a description. We took the top 10 names for each description, names generated from the two baselines (described below) and compared the approaches through an MTurk study.
	The code, data, examples and results are available at this link.\footnote{http://www.cicling.org/2017/data/326}

	\subsection{Baselines}
	\label{ssec:priorart}
	\textbf{Prior Art:}
	{\"O}zbal and Strapparava \cite{ozbal:12}	describe a method to generate names based on homophonic puns and metaphors by combining natural language processing techniques with various linguistic resources available online. 
	We replicated the work of {\"O}zbal and Strapparava \cite{ozbal:12} to generate names. Adapting it our case, the category and properties were provided manually from the descriptions until it output atleast 10 names for the description.
	After following the original specification, if the number of names generated by this algorithm were fewer than the number required for our experimental setup, we added related properties to the set of properties taken as input. For more details about the approach see  \cite{ozbal:12}.
	The output generated from this system was used for further experiments. 
	
	\noindent \textbf{Human:} 10 participants were recruited to give 10 names for one of the descriptions in 4 minutes (time taken by our method). The participants were given information about the criteria being used for creating new names, i.e. unique and appealing names. This experiment gave 10 human generated names for each description. 
	
	\subsection{MTurk Survey: Results and Observations}
	\label{ssec:survey}
	For each description, we created 2 lists of 15 names, each containing 5 names randomly picked from the list of 10 names generated by the three approaches. 
	Table \ref{examplesMethods} shows a few example inputs to the three approaches and the names generated by them. 
	Then, 100 recruited judges from Amazon Mechanical Turk were shown one of the 20 lists and asked to rate each of the 15 names in it as \emph{Good}, \emph{Fair} or \emph{Bad} based on their relevance to the description and uniqueness. Some of the participants of human experiment in Section \ref{ssec:priorart} provided names of currently existing companies. Therefore, the latter instruction was added explicitly to avoid participants rating irrelevant existing names as \emph{Good}. Each list was annotated by 5 judges resulting in 1500 responses.
	
	    \begin{table*}[t]
	        \centering
	        \caption{\label{examplesMethods} Input and Output by the three approaches}

	        \begin{tabular}{lll}
	            \hline \bf Approach & \bf Input & \bf Output \\ \hline
	            Our Method & \begin{tabular}{@{}l@{}} Fabulous furniture to\\ decorate your home \end{tabular} & \begin{tabular}{@{}l@{}}FabFur, MythRate, DressHouse \\ HomeDec, FurDeck \end{tabular}  \\ \hline
	            
	            Our Method & \begin{tabular}{@{}l@{}} Light weight software\\ to locate virus on computer     \end{tabular} & \begin{tabular}{@{}l@{}}FeatherTor, PingWare, CleanDen, \\ FaintCate, ClearSet \end{tabular}  \\ \hline
	            
	            Prior Art & \begin{tabular}{@{}l@{}}Category: furniture; \\ \\ Properties: fabulous, decorative, \\ attractive, homely, comfortable\end{tabular} & \begin{tabular}{@{}l@{}}Woodroom, Houly, Flooroom, \\ Dinnel, Bedroose\end{tabular} \\ \hline
	            
	            Prior Art & \begin{tabular}{@{}l@{}}Category: software; \\ \\Properties: light, locate\\ computable, buggy, safety\end{tabular} & \begin{tabular}{@{}l@{}}Luggyte, Cebuter, Safetyre,\\ Locatr, Coftwarele \end{tabular} \\ \hline
	            
	            Human & \begin{tabular}{@{}l@{}} Fabulous furniture to\\ decorate your home \end{tabular} & \begin{tabular}{@{}l@{}}Decorature, FabHomes, HomeDecor\\ FabulousHomes, FabFurnish\end{tabular}  \\ \hline
	            
	            Human & \begin{tabular}{@{}l@{}} Light weight software\\ to locate virus on computer     \end{tabular} & \begin{tabular}{@{}l@{}}Ubuntu, Nortun, Ad-Blocker,\\  Windows, Web-sites\end{tabular}  \\
	            \hline
	        \end{tabular}
	    \end{table*}
	
	
	\begin{table}
		\small
		\centering
		\caption{\label{resulttable} Ratings for Generated Names}
		\begin{tabular}{|l|l|l|l|}
			\hline \bf Approach & \bf Good & \bf Fair & \bf Bad \\ \hline
			Our Method & 16.6\% & 41.8\% & 41.6\% \\
			Human & 20.4\% & 32.8\% & 46.8\% \\
			Prior Art & 13.2\% & 38.8\% & 48\% \\
			\hline
		\end{tabular}
	\end{table}
	
	%
	%
	
	Table \ref{resulttable} shows percentages of ratings received by names generated by the three approaches
	considering all 10 descriptions. Our method outperforms the prior art. 16.6\% of the names generated by our method received \emph{Good} rating in comparison to 13.2\% of the prior art. Similar is the case in \emph{Fair} rating as well.
	Humans outperform both the automated approaches considering the \emph{Good} rating. However, our method has significantly fewer \emph{Bad} ratings when compared to humans. 
	
	We believe that 4 minutes constraint on humans is harsh. They can think of better names if given sufficient time. As a useful observation, one of the participants seemed to be following our approach to generate names using only the root words. The given description was \emph{Showroom of Fabulous Furniture for Decorating Home}. Our method output names like \emph{HomeDec} and \emph{FabFurNi} which were rated mostly as \emph{Fair} whereas the participant generated names like \emph{HomeDecor} and \emph{FabFurnish} which were rated mostly as \emph{Good}.
	This tells us that the method described in this work is indeed a mechanism that humans use to generate names and further, it can also be used for ideation purposes.

	Additionally, we calculated nDCG \cite{jarvelin2002cumulated}
	to know whether our method's rankings match \emph{Good/Fair/Bad} ratings by human judges. 
	In order to define relevance of name $n$ in nDCG formulation, we used $1$, $0.5$ and $0$ as weights for number of \emph{Good}, \emph{Fair} and \emph{Bad} ratings respectively. 
	The nDCG averaged over 10 descriptions was 0.78 indicating that the ranking generated by our method indeed concurs with human rankings.

	%
	%
	%

\section{Limitations and Future Work}
\label{sec:limitations}

Our methodology will generate homogeneous names given the same description.
Hence, there are opportunities for leveraging enterprise based personalization. 
Further, we agree that there are brand names like Apple, Fox, etc. which cannot be generated by our approach. However, examples like CarMax, DocuSign, etc. and aforementioned online generators, led us to believe that our approach is one of the ways by which humans create names. In future, we plan to investigate abbreviations, reduplications, and modifications over blended syllables to generate better names.

\section{Conclusions}
\label{sec:conclusions}

Our work is one of the first approaches to algorithmically generate appealing brand names from description. In addition to being directly used, the recommended names can also aid in ideation. Quantitative definitions of pronounciability, memorability, and uniqueness have been proposed.
Further, the set of names generated by us is diverse. The inclusion of diversity aids in the ideation process, providing a rich set of names to any user of the system. Achieving near human results certainly opens the door for automation in this human dominated domain.
\section*{Acknowledgments}

We thank Dr. Niloy Ganguly for providing valuable comments and feedback.

\bibliographystyle{splncs}
\bibliography{paper}

\begin{thebibliography}{10}

\bibitem{Ries:81}
Ries, A., Trout, J.:
\newblock Positioning: The battle for your mind (1981)

\bibitem{Robertson:89}
Robertson, K.:
\newblock Strategically desirable brand name characteristics.
\newblock Journal of Consumer Marketing \textbf{6} (1989)  61--71

\bibitem{Yorkston:04}
Yorkston, E., Menon, G.:
\newblock A sound idea: Phonetic effects of brand names on consumer judgments.
\newblock Journal of Consumer Research \textbf{31} (2004)  43--51

\bibitem{Little:10}
Little, G., Chilton, L.B., Goldman, M., Miller, R.C.:
\newblock Exploring iterative and parallel human computation processes.
\newblock In: Proceedings of the ACM SIGKDD workshop on human computation, ACM
  (2010)  68--76

\bibitem{Clements:15}
Clements, J.:
\newblock Generating 56-bit passwords using markov models (and charles
  dickens).
\newblock arXiv preprint arXiv:1502.07786 (2015)

\bibitem{Allbery:88}
Allbery, B.:
\newblock pwgen—-random but pronounceable password generator.
\newblock USENET posting in comp. sources. misc (1988)

\bibitem{Crawford:08}
Crawford, H., Aycock, J.:
\newblock Kwyjibo: Automatic domain name generation.
\newblock Softw. Pract. Exper. \textbf{38} (2008)  1561--1567

\bibitem{Bauer:83}
Bauer, L.:
\newblock English word-formation.
\newblock Cambridge university press (1983)

\bibitem{kondrak2003phonetic}
Kondrak, G.:
\newblock Phonetic alignment and similarity.
\newblock Computers and the Humanities \textbf{37} (2003)  273--291

\bibitem{hedlund2005natural}
Hedlund, G.J., Maddocks, K., Rose, Y., Wareham, T.:
\newblock Natural language syllable alignment: From conception to
  implementation.
\newblock In: Proceedings of the Fifteenth Annual Newfoundland Electrical and
  Computer Engineering Conference. (2005)

\bibitem{ozbal:12}
{\"O}zbal, G., Strapparava, C.:
\newblock A computational approach to the automation of creative naming.
\newblock In: Proceedings of the 50th Annual Meeting of the Association for
  Computational Linguistics: Long Papers-Volume 1, Association for
  Computational Linguistics (2012)  703--711

\bibitem{ozbal:12:brand}
{\"O}zbal, G., Strapparava, C., Guerini, M.:
\newblock Brand pitt: A corpus to explore the art of naming.
\newblock In: Proceedings of the eighth international conference on Language
  Resources and Evaluation (LREC-2012), Istanbul, Turkey, May, Citeseer (2012)

\bibitem{Bird:09}
Bird, S., Klein, E., Loper, E.:
\newblock Natural language processing with Python.
\newblock " O'Reilly Media, Inc." (2009)

\bibitem{Toutanova:03}
Toutanova, K., Klein, D., Manning, C.D., Singer, Y.:
\newblock Feature-rich part-of-speech tagging with a cyclic dependency network.
\newblock In: Proceedings of the 2003 Conference of the North American Chapter
  of the Association for Computational Linguistics on Human Language
  Technology-Volume 1, Association for Computational Linguistics (2003)
  173--180

\bibitem{Wordnet:95}
Miller, G.A.:
\newblock Wordnet: a lexical database for english.
\newblock Communications of the ACM \textbf{38} (1995)  39--41

\bibitem{Goldhahn:12}
Goldhahn, D., Eckart, T., Quasthoff, U.:
\newblock Building large monolingual dictionaries at the leipzig corpora
  collection: From 100 to 200 languages.
\newblock In: LREC. (2012)  759--765

\bibitem{Kincaid:75}
Kincaid, J.P., Jr, R.P.F., Rogers, R.L., Chissom, B.S.:
\newblock Derivation of new readability formulas (automated readability index,
  fog count and flesch reading ease formula) for navy enlisted personnel.
\newblock Technical report, DTIC Document (1975)

\bibitem{Schiavoni:14}
Schiavoni, S., Maggi, F., Cavallaro, L., Zanero, S.:
\newblock Phoenix: Dga-based botnet tracking and intelligence.
\newblock In: Detection of intrusions and malware, and vulnerability
  assessment.
\newblock Springer (2014)  192--211

\bibitem{Mirriam:05}
Webster, M.:
\newblock The Official Scrabble Players Dictionary.
\newblock "Springfield" (2005)

\bibitem{chen1996empirical}
Chen, S.F., Goodman, J.:
\newblock An empirical study of smoothing techniques for language modeling.
\newblock In: Proceedings of the 34th annual meeting on Association for
  Computational Linguistics, Association for Computational Linguistics (1996)
  310--318

\bibitem{Mizil:12}
Danescu-Niculescu-Mizil, C., Cheng, J., Kleinberg, J., Lee, L.:
\newblock You had me at hello: How phrasing affects memorability.
\newblock In: Proceedings of the 50th Annual Meeting of the Association for
  Computational Linguistics: Long Papers-Volume 1, Association for
  Computational Linguistics (2012)  892--901

\bibitem{Michel:11}
Michel, J.B., Shen, Y.K., Aiden, A.P., Veres, A., Gray, M.K., Pickett, J.P.,
  Hoiberg, D., Clancy, D., Norvig, P., Orwant, J.,  et~al.:
\newblock Quantitative analysis of culture using millions of digitized books.
\newblock science \textbf{331} (2011)  176--182

\bibitem{Joachims:02}
Joachims, T.:
\newblock Optimizing search engines using clickthrough data.
\newblock In: Proceedings of the eighth ACM SIGKDD international conference on
  Knowledge discovery and data mining, ACM (2002)  133--142

\bibitem{carbonell1998use}
Carbonell, J., Goldstein, J.:
\newblock The use of mmr, diversity-based reranking for reordering documents
  and producing summaries.
\newblock In: Proceedings of the 21st annual international ACM SIGIR conference
  on Research and development in information retrieval, ACM (1998)  335--336

\bibitem{modani2015creating}
Modani, N., Khabiri, E., Srinivasan, H., Caverlee, J.:
\newblock Creating diverse product review summaries: A graph approach.
\newblock In: International Conference on Web Information Systems Engineering,
  Springer (2015)  169--184

\bibitem{jarvelin2002cumulated}
J{\"a}rvelin, K., Kek{\"a}l{\"a}inen, J.:
\newblock Cumulated gain-based evaluation of ir techniques.
\newblock ACM Transactions on Information Systems (TOIS) \textbf{20} (2002)
  422--446

\end{thebibliography}
\end{document}